\documentclass{article}
\usepackage{spconf}
\usepackage{amsmath,amssymb}
\usepackage{booktabs}
\usepackage{graphicx}
\usepackage{multirow}
\usepackage{array}
\usepackage{xcolor}
\usepackage[hidelinks]{hyperref}
\usepackage{microtype}
\usepackage{subcaption}

\def\tabref#1{Table \ref{#1}}
\def\figref#1{Fig. \ref{#1}}
\def\secref#1{Section \ref{#1}}
\def\eqref#1{Eq (\ref{#1})}
\def\Bdma#1{\mbox{\boldmath{$#1$}}}
\newcommand{\etal}{\textit{et al}.~}

\begin{document}

\title{I-Parakeet: Integer-Only Conformer ASR on Mobile NPU}
\name{Taichi Nishimura}
\address{Sony Interactive Entertainment, Tokyo, Japan}

\maketitle

\begin{abstract}
In this paper, we propose I-Parakeet, an integer-only implementation of NVIDIA's Parakeet-CTC (0.6B parameters) that runs on a smartphone NPU without any floating-point operator or CPU fallback. Modern Conformer ASR models are hard to deploy on edge devices because of their size, and quantized models still fall back to floating point for numerically sensitive operations. This prevents them from fully exploiting integer accelerators such as mobile NPUs. To achieve this, our contributions are threefold. First, we derive an integer formulation of the relative-positional self-attention at the core of the Conformer.
We fuse its two score branches with different quantization scales and the relative shift into integer-only operations. Second, we introduce a minimax-optimized Swish approximation that minimizes the maximum error of the Swish output. Third, a layer-wise range analysis of activations yields two targeted remedies: an INT16 grid for the BatchNorm output and percentile calibration for the heavy-tailed pre-encoder activations. I-Parakeet achieves 4.97\% WER on LibriSpeech test-other, running on a Qualcomm NPU at a real-time factor of 0.048, 7.5$\times$ faster than a CPU baseline.
\end{abstract}


\section{Introduction}
Automatic speech recognition (ASR) has achieved remarkable reductions in word error rate (WER) driven largely by Transformer-based architectures \cite{vaswani2017neurips,radford2023whsiper,wav2vec,peng2024owsm,anmol2020interspeech}.
In particular, Conformer \cite{anmol2020interspeech}, which combines self-attention with convolution to capture both global and local dependencies, has become a de-facto standard.
However, these gains have come through scale.
For example, NVIDIA's Parakeet-CTC \cite{dima2023asru}, a representative modern Conformer, has 0.6B parameters.
At this scale, the weights alone exceed 2.4 GB in single precision, and floating-point inference strains the memory, latency, and power of edge devices, making on-device deployment challenging.

Quantization \cite{gholami2021arxiv} addresses this deployment gap.
INT8 weights reduce the model size of Parakeet-CTC 0.6B from 2.4GB to 600MB.
Most importantly for edge deployment, modern mobile SoCs ship with neural processing units (NPUs) built for integer arithmetic: quantized inference can be offloaded entirely from the CPU/GPU, freeing them for other workloads while reducing power draw.

However, most quantized ASR models are not integer-only. The INT8 weights are dequantized to FP16/FP32 to stabilize numerically sensitive operations such as normalization, softmax, and nonlinear activations. Such models cannot fully exploit NPUs and cannot run at all on processors that lack floating-point units, such as ARM Cortex-M. Kim \etal addressed this for language models with I-BERT \cite{kim2021icml}, designing integer-only kernels for LayerNorm \cite{ba2016arxiv}, Softmax, and GELU \cite{dan2016arxiv}, and later extended the approach to ASR \cite{kim2022icassp}, demonstrating integer-only inference for CNN-\cite{kriman2020icassp} and Conformer-based models \cite{anmol2020interspeech}. However, the hardware evidence across integer-only studies \cite{kim2021icml, kim2022icassp} is limited to GPUs, and whether an integer-only Conformer actually runs on a mobile NPU remains unexplored.

In this paper, we propose I-Parakeet, an integer-only implementation of NVIDIA's Parakeet-CTC that runs on the NPU of a consumer smartphone. We use integer-only as a deployment property: no floating-point operator and no CPU fallback. To deploy a modern Conformer on a real NPU, we resolve three issues.
First, the relative-positional self-attention at the core of the Conformer has not been explicitly formulated for integer-only inference. Its two score branches have different quantization scales, and the relative shift interleaves them with zero-padding. We derive an integer-only formulation that fuses the two branches and the relative shift (\secref{ssec:relpos}).
Second, the Swish activations must be approximated in integer arithmetic. Prior work \cite{kim2021icml} fits the coefficients to the sigmoid or tanh rather than to the Swish output. We instead minimize the maximum error of the Swish output and show that this criterion matters for WER (\secref{ssec:swish}).
Third, several tensors exhibit activation ranges that INT8 cannot cover. Although naive INT8 quantization degrades Parakeet-CTC significantly, moving every tensor to a wider grid harms the NPU's throughput. A layer-wise range analysis assigns an INT16 grid to the BatchNorm output and percentile calibration to the pre-encoder, while other tensors stay INT8 (\secref{ssec:islands}).

\section{I-Parakeet Overview}

\begin{figure}[t]
  \centering
  \begin{subfigure}{\linewidth}
    \centering
    \includegraphics[width=0.9\linewidth]{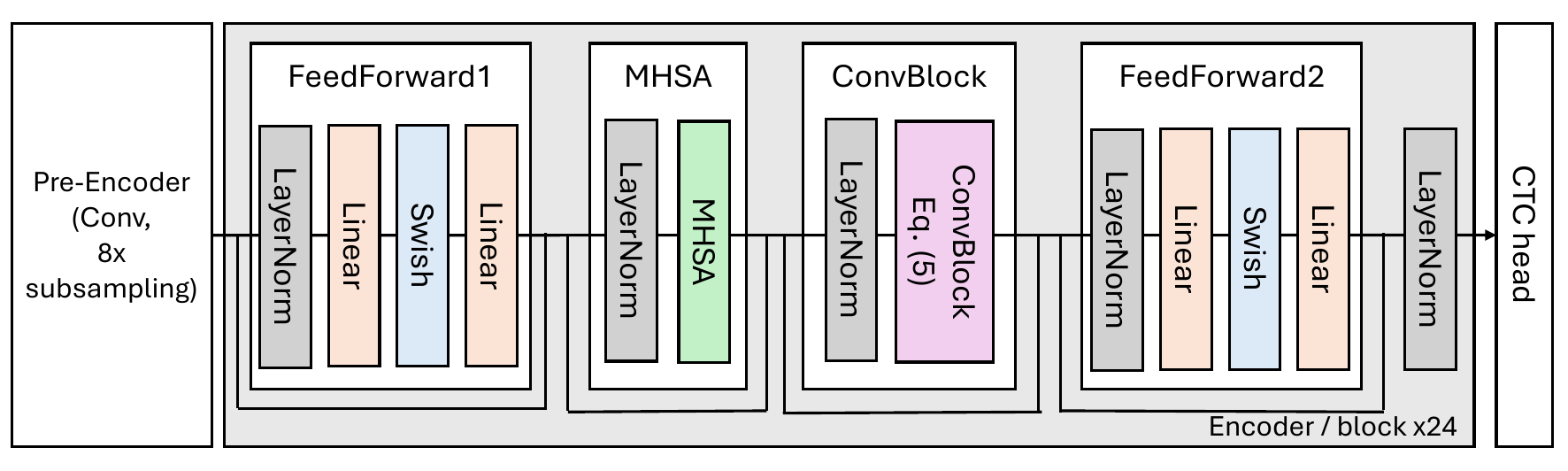}
    \caption{Parakeet-CTC-0.6B}
    \label{fig:overview_fp}
  \end{subfigure}\\[4pt]
  \begin{subfigure}{\linewidth}
    \centering
    \includegraphics[width=0.9\linewidth]{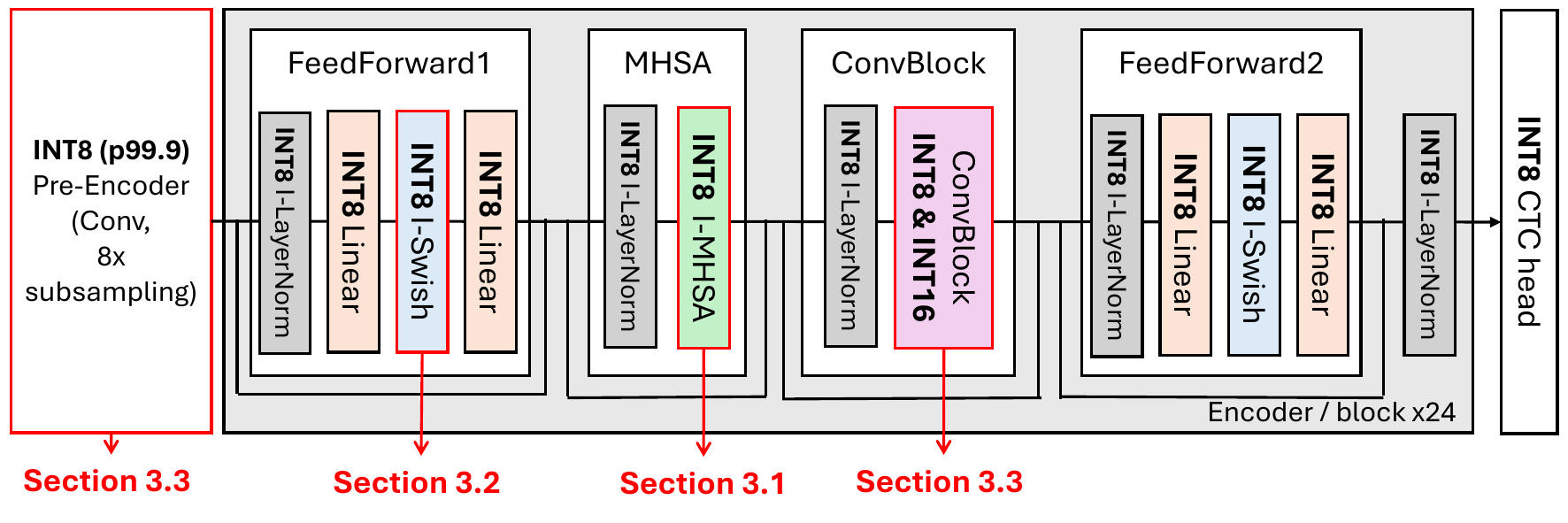}
    \caption{Ours: I-Parakeet-CTC-0.6B}
    \label{fig:overview_int}
  \end{subfigure}
  \caption{Model Overview. The integer relative-position self-attention, integer Swish, and quantization strategy in the pre-encoder/ConvBlock are described in \secref{ssec:relpos}, \ref{ssec:swish}, and \ref{ssec:islands}.}
  \label{fig:overview}
\end{figure}

\subsection{Parakeet-CTC}
\label{ssec:parakeet}

\figref{fig:overview} (a) shows an overview of Parakeet-CTC-0.6B \cite{parakeet}, a Conformer-based CTC model consisting of a convolutional pre-encoder, a stack of $N{=}24$ Conformer blocks (encoder), and a linear CTC head.
Given the input log-mel spectrogram $\Bdma{M} \in \mathbb{R}^{80 \times T_0}$ ($T_0$ is the number of input frames), the pre-encoder firstly reduces its frame rate by a factor of $8$ through three strided convolutions with ReLU activations. Then it projects the result to the model dimension $d=1024$ with a linear layer.
This yields a feature sequence $\Bdma{H}^{(0)} = [\Bdma{h}_1, \dots, \Bdma{h}_L]^\top \in \mathbb{R}^{L \times d}$, where $L = \lceil T_0 / 8 \rceil$ and $\Bdma{h}_l \in \mathbb{R}^{d}$ is the feature vector of the $l$-th frame.

To map $\Bdma{H}^{(0)}$ to a representation suitable for CTC decoding, the encoder applies a stack of conformer blocks.
Each block transforms $\Bdma{X} = \Bdma{H}^{(i-1)}$ into $\Bdma{H}^{(i)}$ through four residual sub-modules:
\begin{align}
  \Bdma{Z}_1 &= \Bdma{X} + 0.5\,\mathrm{FFN}_1\!\left(\mathrm{LN}(\Bdma{X})\right), \label{eq:ffn1} \\
  \Bdma{Z}_2 &= \Bdma{Z}_1 + \mathrm{MHSA}\!\left(\mathrm{LN}(\Bdma{Z}_1)\right), \label{eq:mhsa} \\
  \Bdma{Z}_3 &= \Bdma{Z}_2 + \mathrm{Conv}\!\left(\mathrm{LN}(\Bdma{Z}_2)\right), \label{eq:conv} \\
  \Bdma{H}^{(i)} &= \mathrm{LN}\!\left(\Bdma{Z}_3 + 0.5\,\mathrm{FFN}_2\!\left(\mathrm{LN}(\Bdma{Z}_3)\right)\right), \label{eq:ffn2}
\end{align}
where the ConvBlock $\mathrm{Conv}$ in \eqref{eq:conv} is a five-stage pipeline:
\begin{equation}
  \mathrm{Conv}(\Bdma{U}) = \mathrm{PW}_2\!\left(\mathrm{Swish}\!\left(\mathrm{BN}\!\left(\mathrm{DW}\!\left(\mathrm{GLU}\!\left(\mathrm{PW}_1(\Bdma{U})\right)\right)\right)\right)\right), \label{eq:convmod}
\end{equation}
with $\mathrm{PW}_{1,2}$ pointwise convolutions, $\mathrm{GLU}$ the gated linear unit, $\mathrm{DW}$ a depthwise convolution ($K{=}9$), and $\mathrm{BN}$ BatchNorm. In \eqref{eq:ffn1}--\eqref{eq:ffn2}, $\mathrm{LN}$ denotes LayerNorm, $\mathrm{FFN}_{1,2}$ are feed-forward networks with Swish activations \cite{ramachandran2018iclr}, and $\mathrm{MHSA}$ is multi-head self-attention with relative positional encoding \cite{dai2019acl}. Finally, the CTC head projects each frame to a vocabulary size and greedy decoding produces the transcription.

\subsection{Quantization of Parakeet-CTC}
\label{ssec:iparakeet}
\figref{fig:overview}(b) shows I-Parakeet, which runs the entire network of Parakeet-CTC in integer arithmetic.
Every tensor uses uniform symmetric quantization. A real tensor $\Bdma{x}$ is mapped to its $b$-bit integer tensor $\Bdma{q}_x$ as
\begin{equation}
  \Bdma{q}_x = \big\lfloor \mathrm{clip}(\Bdma{x}, -\alpha_x, \alpha_x)\,/\,S_x \big\rceil,
  \quad S_x = \alpha_x / (2^{b-1}-1).
  \label{eq:quant}
\end{equation}
The clipping range $\alpha_x$ is a scalar computed offline from calibration data (LibriSpeech dev-other). The common choice is the min--max range $\alpha_x = \max|\Bdma{x}|$ over the calibration set, which covers every observed value, but a few outliers spread the range and degrade the rounding precision. A percentile of $|\Bdma{x}|$ (e.g., 99.9\%) is a robust alternative that clips the tail instead. The choice of $\alpha_x$ significantly affects WER and is discussed in \secref{ssec:islands}.
While activations use one $\alpha_x$ per tensor, weights have $\alpha_x$ per channel, thus \eqref{eq:quant} is applied independently channel by channel. The bit width $b$ is $8$ unless stated otherwise.
A linear layer $\Bdma{y} = \Bdma{W}\Bdma{x}$ then needs only integer operations. The product $\Bdma{q}_W \Bdma{q}_x$ accumulates in INT32 and is requantized to the grid of $\Bdma{y}$:
\begin{equation}
  \Bdma{q}_y = \Big\lfloor \frac{S_W S_x}{S_y}\,\Bdma{q}_W \Bdma{q}_x \Big\rceil
  = \big\lfloor 2^{-n}\, m\,\Bdma{q}_W \Bdma{q}_x \big\rceil,
  \label{eq:requant}
\end{equation}
where the fixed-point multiplier $m = \big\lfloor 2^{n} S_W S_x / S_y \big\rceil$ is precomputed offline ($n{=}16$). Since the real-valued ratio $S_W S_x / S_y$ never appears at run time, a linear layer reduces to an integer matrix product, an integer multiplication by $m$, and a right $n$ bit shift, with no floating-point operation and no division \cite{fixedpoint}. This is what we call integer-only arithmetic. Convolutions are linear as well and follow the same rule. BatchNorm is folded into the preceding depthwise convolution. For LayerNorm and Softmax we use the integer kernels proposed in I-BERT \cite{kim2021icml}.

To realize I-Parakeet, this paper addresses three issues that are not treated in the previous work. First, we derive an integer relative-positional MHSA, whose scores are the sum of two products with different scales and cannot be requantized by \eqref{eq:requant} directly (\secref{ssec:relpos}). Second, we propose an integer Swish approximation and show that minimizing the maximum error of the Swish output matters for WER (\secref{ssec:swish}, \ref{ssec:accuracy}).
Third, a layer-wise analysis of activations finds a few tensors whose range a single INT8 grid cannot cover: the BatchNorm output and the pre-encoder activations. We spread the grid from $b{=}8$ to $16$ for the former and squeeze $\alpha_x$ to the 99.9th percentile for the latter, while all other tensors stay INT8 (\secref{ssec:islands}).

\section{Method}
\label{sec:method}

\subsection{Integer Relative-Positional Self-Attention}
\label{ssec:relpos}

In this section, we derive an integer-only formulation of the relative-positional MHSA in Parakeet.
We first describe its floating-point definition.
For one head, let $\Bdma{Q}, \Bdma{K}, \Bdma{V} \in \mathbb{R}^{L \times d_k}$ be the query, key, and value obtained by linear projections of the block input, and let $\Bdma{P} \in \mathbb{R}^{(2L-1) \times d_k}$ be the projection of the sinusoidal relative-position embeddings. With learned biases $\Bdma{u}, \Bdma{v} \in \mathbb{R}^{d_k}$, the attention scores decompose into a content branch $\Bdma{A}^{\mathrm{c}}$ and position branch $\Bdma{A}^{\mathrm{p}}$:
\begin{align}
  \Bdma{A}^{\mathrm{c}} &= (\Bdma{Q} + \Bdma{1}\Bdma{u}^\top)\,\Bdma{K}^\top,\Bdma{A}^{\mathrm{p}} = \Phi\!\left((\Bdma{Q} + \Bdma{1}\Bdma{v}^\top)\,\Bdma{P}^\top\right), \label{eq:bd} \\
  \Bdma{A} &= \mathrm{softmax}\!\left(\left(\Bdma{A}^{\mathrm{c}} + \Bdma{A}^{\mathrm{p}}\right) / \sqrt{d_k}\right), \Bdma{O}=\Bdma{A}\Bdma{V}
  \label{eq:score}
\end{align}
where $\Phi$ is the relative-shift operation that realigns the $(2L{-}1)$ relative positions to the $L$ absolute key positions.
An integer-only implementation must fix one INT8 grid per tensor at compile time.
The two branches are calibrated independently, $\Bdma{A}^{\mathrm{c}} \approx S_{\mathrm{c}}\,\Bdma{q}_{\mathrm{c}}$ and $\Bdma{A}^{\mathrm{p}} \approx S_{\mathrm{p}}\,\Bdma{q}_{\mathrm{p}}$ with $S_{\mathrm{c}} = S_Q S_K$ and $S_{\mathrm{p}} = S_Q S_P$. Since $S_{\mathrm{c}} \neq S_{\mathrm{p}}$ in general, the sum $\Bdma{q}_{\mathrm{c}} + \Bdma{q}_{\mathrm{p}}$ does not represent $\Bdma{A}^{\mathrm{c}} + \Bdma{A}^{\mathrm{p}}$. Therefore, the two tensors must be converted to a common grid before the addition in \eqref{eq:score}.

To resolve this, we fuse the conversion, addition, and the $1/\sqrt{d_k}$ scaling into a single requantization. Given the grid of fused scores $S_{\mathrm{s}}$, the fused integer scores are computed as:
\begin{align}
  \Bdma{q}_{\mathrm{s}}
  &= \left\lfloor
      \frac{S_{\mathrm{c}}}{S_{\mathrm{s}}\sqrt{d_k}}\,\Bdma{q}_{\mathrm{c}}
    + \frac{S_{\mathrm{p}}}{S_{\mathrm{s}}\sqrt{d_k}}\,\Phi(\Bdma{q}_{\mathrm{p}})
    \right\rceil \label{eq:fusion} \\
  &= \left\lfloor 2^{-n}\!\left(m_{\mathrm{c}}\,\Bdma{q}_{\mathrm{c}} + m_{\mathrm{p}}\,\Phi(\Bdma{q}_{\mathrm{p}})\right) \right\rceil,
  \label{eq:fusion_fx}
\end{align}
where $m_{\mathrm{c}} = \big\lfloor 2^{n}\,S_{\mathrm{c}}/(S_{\mathrm{s}}\sqrt{d_k}) \big\rceil$ and $m_{\mathrm{p}} = \big\lfloor 2^{n}\,S_{\mathrm{p}}/(S_{\mathrm{s}}\sqrt{d_k}) \big\rceil$. The softmax and the product with $\Bdma{V}$ follow I-BERT and \eqref{eq:requant}, so the whole MHSA runs in integer arithmetic.
For efficient inference, we note two points on the position branch. First, $\Bdma{P}$ depends only on the sequence length, so we quantize it offline and store $\Bdma{q}_P$ as an INT8 constant. Second, $\Phi$ only moves values and inserts zeros without any arithmetic. Since moving values does not change them, $\Phi$ gives the same result before and after quantization: $\Phi(S\,\Bdma{q}) = S\,\Phi(\Bdma{q})$. Hence, we apply $\Phi$ to the integer tensor $\Bdma{q}_{\mathrm{p}}$ directly, which we implement as a static index map.

\subsection{Minimax-Optimized Integer Swish}
\label{ssec:swish}

Parakeet-CTC has the Swish activation $\mathrm{sw}(x) = x\,\sigma(x)$ in every feed-forward and ConvBlock layer, and the sigmoid $\sigma$ has no integer kernel.
As with I-BERT \cite{kim2021icml}, we approximate $\sigma$ with a second-order polynomial. Since $\sigma(x) = (1+\tanh(x/2))/2$, we approximate $\tanh$, which is an odd function, thus is enough to fit the polynomial on $u \ge 0$ and mirror it to $u<0$.
Following previous work, we use the polynomial $a(u - c)^2 + 1$, whose vertex at $u = c$ takes the value $1$ (the saturated value of $\tanh$) and clip the output to $1$ for $u > c$:
\begin{equation}
  \widehat{\tanh}(u) = \mathrm{sgn}(u)\left[a\left(\min(|u|, c) - c\right)^2 + 1\right].
  \label{eq:itanh}
\end{equation}
I-BERT adopts least-squares fit to optimize $(a, c)$ to minimize the mean squared error ($L_2$) between $\widehat{\tanh}$ and $\tanh$.

However, we found that this approach is sub-optimal for Parakeet-CTC due to two issues.
First, the $L_2$ treats all $x$ equally, whether $x$ is near $0$ or far from it, but in Swish, the $\sigma$ is multiplied by $x$, causing the big error on large $|x|$ (e.g., The same $\sigma$ error is 40 times larger at $x=4$ than at $x=0.1$). Based on this, we assume that minimizing the maximum error ($L_\infty$) over $x$ is better than the average $L_2$.
Second, although I-BERT fits the $\tanh$, the coefficients that best fit the $\tanh$ are not necessarily the ones that best fit the Swish output. To this end, we focus on fitting the Swish output directly to reduce the error at the output.

Based on them, we optimize $(a, c)$ that minimizes the maximum error of the Swish output:
\begin{equation}
  (a^{*}, c^{*}) = \arg\min_{a,\,c}\ \max_{x}\ \left|\mathrm{sw}(x) - \widehat{\mathrm{sw}}(x;\,a, c)\right|,
  \label{eq:minimax}
\end{equation}
where $\widehat{\mathrm{sw}}(x;\,a, c) = x\,(1+\widehat{\tanh}(x/2))/2$. We solve \eqref{eq:minimax} numerically and obtain $a^{*} = -0.1240$ and $c^{*} = 2.4632$.

\subsection{Layer-wise Activation Range Analysis}
\label{ssec:islands}

\begin{figure}[t]
  \centering
  \includegraphics[width=0.8\linewidth]{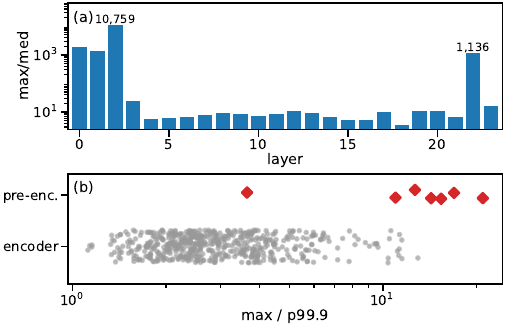}
  \caption{(a) Per-channel spread of the BatchNorm scale $\gamma_c/\sqrt{\sigma_c^2+\epsilon}$ per layer (max/median). (b) Ratio of the observed max to the 99.9 percentile for all calibrated activations.}
  \label{fig:layer_analysis}
\end{figure}

We profile the ranges of all activations over the calibration data and find that the default settings (bit-width $b=8$ and min–max range $\alpha_x$ for every tensor) fails at two places due to outliers \cite{tim2022nips,xiao2023icml}. \figref{fig:layer_analysis} summarizes the analysis. We change $b$ at one place and $\alpha_x$ at the other, and keep the default elsewhere.

The first failure is the BatchNorm output in the ConvBlock.
\figref{fig:layer_analysis} (a) shows the ratio of the max to the median of the BatchNorm coefficient $\gamma_c/\sqrt{\sigma_c^2+\epsilon}$ across channels. The ratio exceeds four orders of magnitude in several layers. A per-tensor INT8 grid must cover the few extreme channels, so the remaining channels are left with almost no quantization levels. We therefore set $b=16$ for the BatchNorm output. In \tabref{tab:ablation} we call this setting per-tensor INT16, and the default per-tensor INT8; both names refer to the BatchNorm only.

The second failure is the pre-encoder activations.
\figref{fig:layer_analysis} (b) shows the ratio of the max to the 99.9th percentile of the activation values. In the pre-encoder the ratio reaches $11$--$21\times$, so a min--max $\alpha_x$ spends most of the INT8 grid on a $0.1\%$ tail. Inside the encoder it stays at $2$--$3\times$. Therefore we set $\alpha_x$ to the 99.9th percentile in the pre-encoder and keep min--max in the encoder. In \tabref{tab:ablation} we call this hybrid scaling.

\section{Experiments}
\label{sec:experiments}

\begin{table}[t]
  \centering
  \caption{On-device comparison on LibriSpeech test-other (Nothing Phone (3a)). The unit of peak memory is MiB.}
  \label{tab:ondevice}
  \scalebox{0.71}{
  \begin{tabular}{lccccc}
    \toprule
    Method & Backend & Weights / compute & WER & RTF & Peak memory \\
    \midrule
    parakeet.cpp \cite{parakeetcpp} & CPU & FP16 / FP32 & 3.76 & 0.36 & 1517 \\
    parakeet.cpp \cite{parakeetcpp} & CPU & INT8 / FP32 & 3.76 & 0.42 & 1045 \\
    \midrule
    Stock FP16 & NPU & FP16 / FP16 & 100.0 & -- & -- \\
    Stock INT8 PTQ & NPU & INT8 / INT8 & 100.0 & -- & -- \\
    I-Parakeet & NPU & INT8 / INT8 \& INT16 & 4.97 & \textbf{0.048} & \textbf{612} \\
    \bottomrule
  \end{tabular}
  }
\end{table}

\subsection{On-Device Results}
\label{ssec:device}
\noindent
\textbf{Settings.} We evaluate WER, RTF, and peak memory on a mid-range consumer smartphone (Nothing Phone (3a), Snapdragon 7s Gen~3, SM7635) with QNN SDK (QAIRT 2.47) as the NPU runtime. Since the NPU requires static shapes, we follow prior work~\cite{lee2026arxiv} and compile graphs for input lengths of 3 to 35 seconds, routing each utterance to the smallest graph that contains it. Padding with silence features adds 23\% to the processed frames, and all RTFs include this overhead. WER is computed with the Whisper English text normalizer.
The baselines are parakeet.cpp~\cite{parakeetcpp} on two CPU threads with FP16 and q8\_0 weights (both with FP32 compute), and the same checkpoint converted by the stock Qualcomm SDK toolchain to FP16 and to INT8 by post-training quantization.

\noindent
\textbf{Results.} \tabref{tab:ondevice} summarizes the results. I-Parakeet runs entirely on the NPU without CPU fallback, 7.5$\times$ faster than FP16 parakeet.cpp with 60\% less peak memory, at a WER cost of 1.21 points. Weight-only INT8 does not speed up the CPU path, so the gain comes from integer compute on the NPU, not from smaller weights.
The on-device WER of 4.97\% is slightly below the 5.32\% of our integer simulator (\secref{ssec:accuracy}) because the NPU evaluates the sigmoid in Swish with a hardware lookup table instead of the polynomial. Both stock paths give 100\% WER: the FP16 failure is consistent with the pre-encoder activations of \figref{fig:layer_analysis} exceeding the FP16 range, and the INT8 failure shows that tool-supported post-training quantization is not sufficient for this model.

\subsection{Accuracy Analysis in Simulation}
\label{ssec:accuracy}

\noindent
\textbf{Settings.} To isolate each design choice, we use a PyTorch implementation of the integer arithmetic that mirrors the deployed graph but has Swish polynomial approximation in \secref{ssec:swish}.
We evaluate on LibriSpeech \cite{panayotov2015icassp} and Common Voice \cite{ardila2020lrec}.
We compare I-Parakeet with two integer-only baselines that share the integer attention of \secref{ssec:relpos}. I-BERT recipe \cite{kim2021icml,kim2022icassp} quantizes every tensor to INT8 with min--max ranges and approximates Swish by the least-squares fit to the tanh. Naive INT8 uses the same INT8 min--max quantization but our minimax Swish, thus the difference between Naive INT8 and I-Parakeet is the two range remedies of \secref{ssec:islands}.

\noindent
\textbf{Results.} \tabref{tab:main} reports WER on both datasets. I-Parakeet reaches 2.61\% on test-clean and 5.32\% on test-other, 0.74/1.56 points behind FP32. Compared with Naive INT8, it recovers 0.40/1.06 points, the contribution of the two range remedies, and the same ordering holds on Common Voice.

\noindent
\textbf{Ablation 1: Swish approximation.} \tabref{tab:ablation} varies one design choice at a time. The ranking of the maximum output error matches the ranking of WER. On the norm perspective, $L_\infty$ beats $L_2$ for both fitting targets. In terms of fitting target function, fitting the Swish output beats fitting the tanh under both norms. Therefore, the two axes contribute independently. Hard-Swish \cite{howard2019mobilenetv3} degrades WER by one point.

\noindent
\textbf{Ablation 2: BatchNorm output precision.} Per-tensor INT8 for the BatchNorm output degrades WER to 5.91. Per-channel INT16 reaches 5.29 but is not supported for activations on the NPU. Our deployable per-tensor INT16 attains 5.32, within 0.03 points of per-channel INT16.

\noindent
\textbf{Ablation 3: Scale calibration.} Hybrid scaling beats min--max everywhere. Applying p99.9 everywhere is far worse (8.03), so clipping helps only where the distribution is heavy-tailed.

\begin{table}[t]
  \centering
  \caption{Simulated WER on PyTorch; on-device results are in \tabref{tab:ondevice}.}
  \label{tab:main}
  \scalebox{0.8}{
  \begin{tabular}{lccc}
    \toprule
    & \multicolumn{2}{c}{LibriSpeech} & Common Voice \\
    \cmidrule(lr){2-3} \cmidrule(lr){4-4}
    Model & test-clean & test-other & test \\
    \midrule
    Parakeet-CTC (FP32) & 1.87 & 3.76 & 10.55 \\
    \midrule
    I-BERT recipe \cite{kim2021icml,kim2022icassp} & 3.41 & 7.41 & 19.05 \\
    Naive INT8 & 3.01 & 6.38 & 16.54 \\
    I-Parakeet & \textbf{2.61} & \textbf{5.32} & \textbf{14.70} \\
    \bottomrule
  \end{tabular}
  }
\end{table}

\begin{table}[t]
  \centering
  \caption{Ablation on LibriSpeech test-other (WER). Max err.\ is the max absolute error of the approximated Swish output.}
  \label{tab:ablation}
  \scalebox{0.8}{
  \begin{tabular}{lcc}
    \toprule
    Configuration & Max err. & WER \\
    \midrule
    \multicolumn{3}{l}{\textit{Swish approximation (Sec.~\ref{ssec:swish})}} \\
    \quad $L_\infty$ fit to Swish (ours) & 0.039 & 5.32 \\
    \quad $L_2$ fit to Swish & 0.045 & 5.49 \\
    \quad $L_\infty$ fit to tanh & 0.068 & 5.81 \\
    \quad $L_2$ fit to tanh & 0.073 & 5.95 \\
    \quad Hard-Swish & 0.142 & 6.33 \\
    \midrule
    \multicolumn{3}{l}{\textit{BatchNorm output precision (Sec.~\ref{ssec:islands})}} \\
    \quad per-tensor INT16 (ours) & -- & 5.32 \\
    \quad per-channel INT16 (not deployable) & -- & 5.29 \\
    \quad per-tensor INT8 (min--max) & -- & 5.91 \\
    \midrule
    \multicolumn{3}{l}{\textit{Scale calibration on pre-encoder and encoder (Sec.~\ref{ssec:islands})}} \\
    \quad hybrid scaling of min--max and p99.9 (ours) & -- & 5.32 \\
    \quad min--max only & -- & 5.64 \\
    \quad p99.9 only & -- & 8.03 \\
    \bottomrule
  \end{tabular}
  }
\end{table}

\section{Conclusion}
\label{sec:conclusion}

We proposed I-Parakeet, an integer-only Parakeet-CTC-0.6B running on a smartphone NPU at an RTF of 0.048 with 4.97\% WER on LibriSpeech test-other, showing that modern Conformer ASR needs no floating-point inference.

\bibliographystyle{IEEEbib}
\bibliography{refs}

\end{document}